\def\year{2019}\relax
\documentclass[letterpaper]{article} %DO NOT CHANGE THIS
\usepackage{aaai19}  %Required
\usepackage{times}  %Required
\usepackage{helvet}  %Required
\usepackage{courier}  %Required
\usepackage{url}  %Required
\usepackage{graphicx}  %Required
\usepackage{mathtools}
\usepackage{amsmath}
\usepackage{color}
\usepackage{booktabs}
\usepackage{hyperref}
\usepackage[usenames,dvipsnames]{xcolor}

\DeclarePairedDelimiter\norm{\lVert}{\rVert}
\usepackage{kotex}
\makeatletter
\renewcommand\@biblabel[1]{}
\makeatother

\begin{document}
% The file aaai.sty is the style file for AAAI Press 
% proceedings, working notes, and technical reports.
%
\title{Knowledge Distillation with Adversarial Samples \\ Supporting Decision Boundary}
% \author{Anonymous author(s)\\
% Paper ID 1496
% % Affilliation\\
% % Address\\
% % Address\\
% }

\author{
  Byeongho Heo\textsuperscript{1}\thanks{Authors contributed equally.}
  \,\,
  Minsik Lee\textsuperscript{2}\footnotemark[1]
  \,\,
  Sangdoo Yun\textsuperscript{3}
  \,\,
  Jin Young Choi\textsuperscript{1}
  \\
  \small{\texttt{\{bhheo, jychoi\}@snu.ac.kr},\; 
  \texttt{mleepaper@hanyang.ac.kr},\;
  \texttt{sangdoo.yun@navercorp.com}}\\
  \textsuperscript{1}Department of ECE, ASRI, Seoul National University, Korea\\
  \textsuperscript{2}Division of EE, Hanyang University, Korea\\
  \textsuperscript{3}Clova AI Research, NAVER Corp, Korea\\
}

\nocopyright
\maketitle
\begin{abstract}
Many recent works on knowledge distillation have provided ways to transfer the knowledge of a trained network for improving the learning process of a new one, but finding a good technique for knowledge distillation is still an open problem.
In this paper, we provide a new perspective based on a decision boundary, which is one of the most important component of a classifier. The generalization performance of a classifier is closely related to the adequacy of its decision boundary, so a good classifier bears a good decision boundary. 
%Therefore, transferring the boundaries directly can be a good attempt for knowledge distillation.
Therefore, transferring information closely related to the decision boundary can be a good attempt for knowledge distillation.
To realize this goal, we utilize an adversarial attack to discover samples supporting a decision boundary. Based on this idea, to transfer more accurate information about the decision boundary, the proposed algorithm trains a student classifier based on the adversarial samples supporting the decision boundary. 
%Alongside, two metrics are proposed to evaluate the similarity between decision boundaries. 
Experiments show that the proposed method indeed improves knowledge distillation and achieves the state-of-the-arts performance.
\footnote[1]{Code is available at \href{https://github.com/bhheo/BSS_distillation}{\url{https://github.com/bhheo/BSS_distillation}}}
\end{abstract}

\section{Introduction}

Knowledge distillation is a method to enhance the training of a new network based on an existing, already trained network.
In a teacher-student framework, the existing network is considered as a teacher and the new network becomes a student.
\citeauthor{Hinton}~\shortcite{Hinton}, a pioneer in knowledge distillation, proposed a loss minimizing the cross-entropy between the outputs of the student and the teacher, which referred to as the knowledge distillation loss (KD loss).
Due to the KD loss, the student network is trained to be a better classifier than the network trained without knowledge distillation.
Although the goals of the knowledge distillation are diverse, recent studies~\cite{Gift_distill,Detect_distill} focus on improving a small student network using a large network as a teacher using a large teacher network.
These studies aim to create a small network with the speed of a small network and the performance of a large network.
This paper, too, focuses on knowledge distillation in the respect of enhancing the performance of a small network using a large network.

Many of recent studies are focusing on manipulating the KD loss for various purposes.
\citeauthor{FITNET}~\shortcite{FITNET} and \citeauthor{Attention}~\shortcite{Attention} proposed new distillation losses to transfer the hidden layer response of the network and used it with the KD loss. 
\citeauthor{Detect_distill}~\shortcite{Detect_distill} and \citeauthor{Face_distill}~\shortcite{Face_distill} designed new distillation losses for other applications based on the KD loss.
In contrast to these existing approaches that concentrate on how to manipulate various parts of a network in order to improve the effect of knowledge distillation, in this paper, we investigate informative samples for an effective knowledge transfer.
In general, samples near the decision boundary of a classifier have a larger impact on the performance than those far apart from it~\cite{SVM}.
Therefore, if we can generate samples close to the decision boundary, the knowledge of a teacher network would be transferred more effectively by utilizing those samples.
 
To obtain the informative samples close to the decision boundary, we utilize an adversarial attack~\cite{GoodFellow,DeepFool}. 
An adversarial attack is a technique to tamper with the result of a classifier by adding a small perturbation to an input image.
Although an adversarial attack is not particularly aimed at finding a decision boundary, they are closely related to each other~\cite{region_classification}. 
An adversarial attack tries to find a small modification that can change the class of a sample, i.e., it tries to move the sample beyond a nearby decision boundary.
Inspired by this fact, we propose to perform knowledge distillation with the help of an adversarial attack.
To get samples beneficial to knowledge distillation, we modify an attack scheme to search an adversarial sample supporting a decision boundary. The resulting sample is referred to as the boundary supporting sample (BSS) in this paper. A new loss function using BSSs is suggested for knowledge distillation that transfers decision boundary to a student classifier. 
In order to verify whether the proposed method actually transfers the decision boundaries, we also propose two similarity metrics that compares the decision boundaries of two classifiers and use these metrics to examine the decision boundaries of a teacher and a student.

\begin{figure*}[t]
\begin{center}
\centerline{\includegraphics[width=1.0\linewidth]{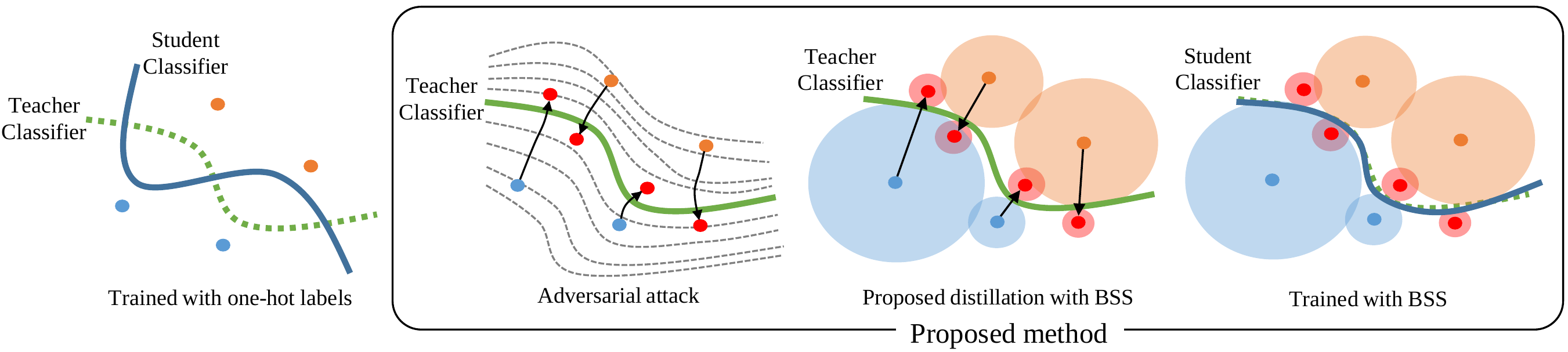}}
\caption{The concept of knowledge distillation using samples close to the decision boundary. The dots in the figure represent the training sample and the circle around a dot represents the distance to the nearest decision boundary. The samples close to the decision boundary enable more accurate knowledge transfer.}
\label{figure1}
\end{center}
\end{figure*}

The proposed method is verified through experiments.
First, we show that the use of BSSs could improve the knowledge distillation scheme of \citeauthor{Hinton}~\shortcite{Hinton} in an image classification problem.
After this, we perform more experiments to examine the generalization performance of the proposed method, of which the result indicates that the proposed method has better generalization performance, and as a result, it can provide good results with less training samples.
Finally, we analyze the proposed method with various experiments.

\subsection{Related Works}
\label{sec:related}
Many studies have been conducted for knowledge distillation since \citeauthor{Hinton}~\shortcite{Hinton} proposed the first knowledge distillation method based on class probability.
\citeauthor{FITNET}~\shortcite{FITNET} used the hidden layer response of a teacher network as a hint for a student network to improve knowledge distillation.
\citeauthor{Attention}~\shortcite{Attention} found the area of activated neurons in a teacher network and transferred the activated area to a student network.
In the case of \citeauthor{Gift_distill}~\shortcite{Gift_distill}, the channel response of a teacher network was transferred for knowledge distillation.
\citeauthor{distill_GAN}~\shortcite{distill_GAN} proposed a knowledge distillation method based on the framework of generative adversarial network~\cite{GAN}.
Some studies~\cite{Detect_distill,Face_distill,distill_similarity} extended knowledge distillation to computer vision applications.
Knowledge distillation has been studied in various directions, but most of the studies are focused on manipulating the hidden layer response of a network or changing the loss appropriately for the purpose.
As far as we know, the proposed method is the first method to improve knowledge distillation by changing the samples used for training.

In the meantime, \citeauthor{Szegedy}~\shortcite{Szegedy} found that a classifier based on a neural network could be fooled easily by a small noise. This work gave rise to a new research topic in neural networks called an adversarial attack, which is about finding a noise that can deceive a neural network.
\citeauthor{DeepFool}~\shortcite{DeepFool} proposed a method to optimize a classifier based on a linear approximation to find the closest adversarial example.
\citeauthor{GoodFellow}~\shortcite{GoodFellow} proposed the adversarial training which trains a classifier with adversarial samples in order to make the network robust to an adversarial attack.
\citeauthor{region_classification}~\shortcite{region_classification} found that an adversarial sample was located near the decision boundary, and used this property to defense an adversarial attack.
There have also been some works that connect an adversarial attack to another research topic.
\citeauthor{DefenseDistillation}~\shortcite{DefenseDistillation} found that a network trained by knowledge distillation is robust to adversarial attacks.
The relationship between an adversarial attack and a decision boundary was used to prevent an adversarial attack in \citeauthor{region_classification}~\shortcite{region_classification}.
Knowledge distillation was also used to prevent an adversarial attacks in \citeauthor{DefenseDistillation}~\shortcite{DefenseDistillation}.
Our study is closely related to these approaches, except that we take an opposite direction to them:
We use adversarial attacks to find decision boundary to enhance knowledge distillation, which is a novel approach that has not been attempted yet.

% \section{Knowledge distillation with adversarial samples}
\section{Method}

\subsection{Adversarial attack for knowledge distillation}
\label{sec:anal}

An adversarial attack is to change a sample in a class into an adversarial sample in another class for a given classifier. In our paper, the given sample for the adversarial attack is referred to as the base sample. In this section, we present an idea to utilize the adversarial attack in knowledge distillation. The idea is about using 
adversarial samples near a decision boundary to transfer the knowledge related with the boundary. 
In the following sections, we first explain the definition of {\it boundary supporting sample} (BSS) and its benefits in knowledge distillation, and then we provide an iterative procedure to find BSSs.

\subsubsection{Benefits of BSSs in knowledge distillation}

It is well-known that the generalization performance of a classifier highly depends on how well the classifier learns the true decision boundary between the actual class distributions~\cite{SVM,bishop2006pattern}. 
This indicates that if a classifier yields a good performance then it probably has good decision boundary that is close to the true one. 
We can analyze knowledge distillation in this respect. 
The knowledge distillation approaches attempt to resolve the generalization issue with help of the trained network with high-performance,
i.e., a teacher, by transferring its knowledge to the classifier we are to train, i.e., the student~\cite{Hinton,FITNET,Attention,Gift_distill}. If we train the student without knowledge distillation, then its performance may not as good as the teacher, which indicates that a decision boundary of the teacher is likely better than that of the student, as shown in Figure~\ref{figure1}. On the other hand, knowledge distillation can enhance the performance of the student, which suggests that the decision boundary of the student is getting improved by knowledge distillation.

However, existing works do not explicitly address that the information about the decision boundary is transferred by knowledge distillation. 
In our paper, inspired by this motivation, we utilize adversarial samples obtained from the training samples to transfer the glimpse of a more accurate decision boundary. 
A boundary supporting sample (BSS) is defined in this respect, it is an adversarial sample that lies near the decision boundary of a teacher classifier.
Since BSSs are labeled samples near decision boundary as depicted in the second picture of Figure \ref{figure1}, they contain the information about the decision boundary. Hence, using BSSs in knowledge distillation can provide a more accurate transfer of decision boundary information.
An BSS in our work is obtained by a gradient descent method based on a classification score functions, and it contains information about both the distance and the path direction from the base sample to the decision boundary.
In conclusion, BSSs could be beneficial to improve the decision boundary, and hence the generalization performance, of a student classifier in knowledge distillation. 

\subsubsection{Iterative Scheme to find a BSS}

\begin{figure}[t]
\begin{center}
\centerline{\includegraphics[width=1\linewidth]{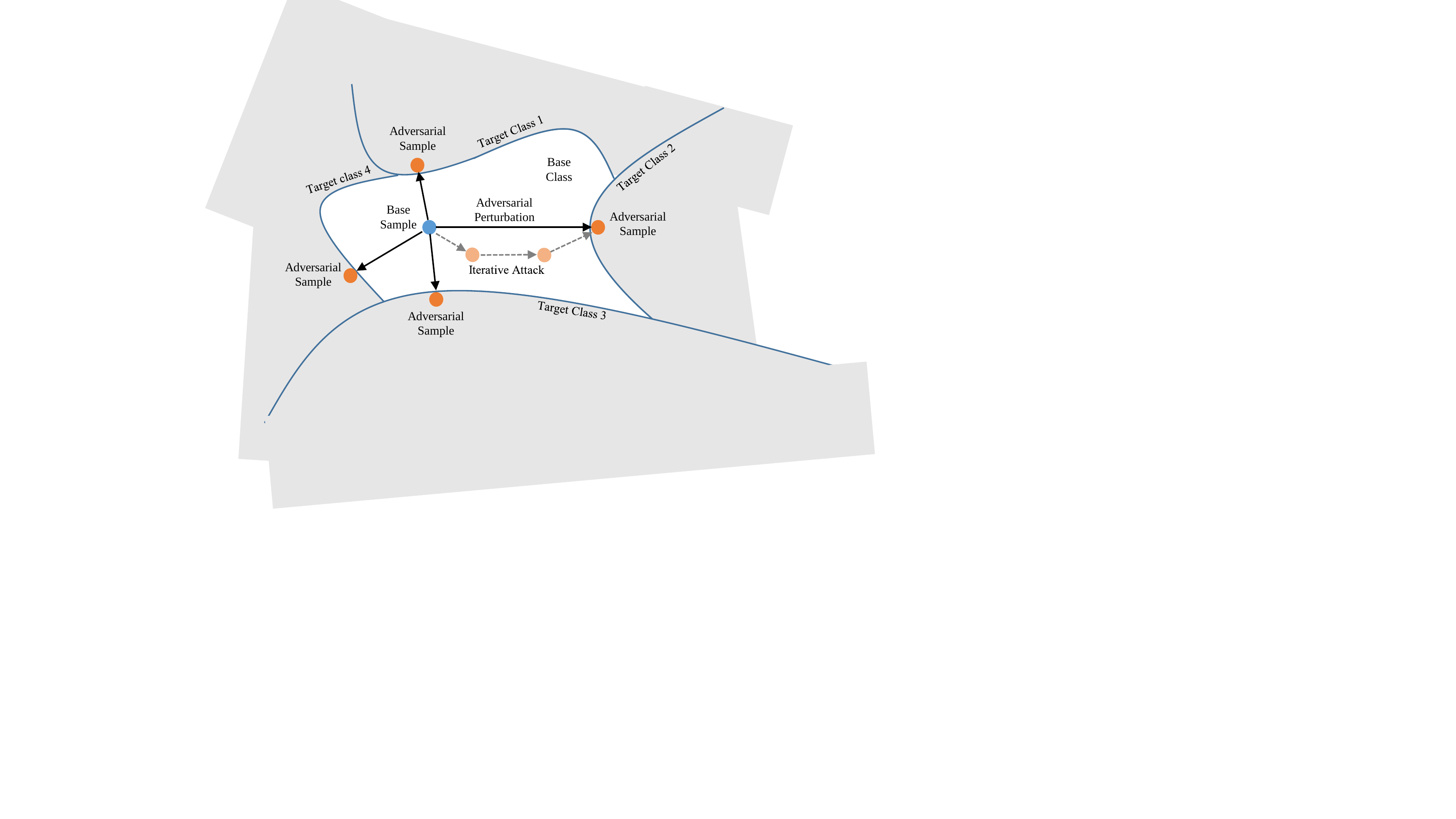}}
\caption{Iterative scheme to find BSSs for a base sample }
\label{figure2}
\end{center}
\end{figure}

For a given sample, as shown in Figure~\ref{figure2}, there exist many BSSs over all classes except the base class that contains the base sample. To find a BSS, we define a loss function based on classification scores produced by a classifier. Then, we search a BSS in the gradient direction of the loss function based on the method in ~\cite{DeepFool} with a modified update rule.

Given a sample vector $\boldsymbol{x}$ in a base class, its corresponding adversarial samples are calculated based on an iterative procedure. First, a sample is initialized to $\boldsymbol{x}^{k}_0 = \boldsymbol{x}$, and then it is iteratively updated to a target class $k$, $k=, 1, 2, \cdots K$. Here, the adversarial sample after the $i$-th iteration is denoted by $\boldsymbol{x}^{k}_i$.
Assume that the classifier $f$ produces classification scores for all classes, where the class of a sample is determined by the class having the maximum score. Then, let $f_{b}(\boldsymbol{x})$ and $f_{k}(\boldsymbol{x})$ be the classification scores for the base class and the target class $k$, respectively.

The goal of the adversarial attack is to decrease the score for the base class while increasing the score for the target class. To this end, the loss function for the attack to the target class $k$ is defined by
\begin{equation}
{\cal L}_k(\boldsymbol{x}) = f_{b} (\boldsymbol{x}) - f_{k}(\boldsymbol{x}).
\label{score_diff}
\end{equation}
This loss becomes zero at a point on the decision boundary, positive at a point within the base class, and negative at an adversarial point within the target class. To find an adversarial sample, we move the sample to the direction minimizing the loss by the iterative scheme in (\ref{iteration}), until the loss becomes negative.
\begin{equation}
\boldsymbol{x}^{k}_{i+1} = \boldsymbol{x}^{k}_{i} - \eta \left ( {\cal L}_k(\boldsymbol{x}^{k}_{i})+\epsilon \right) \frac{\nabla{\cal L}_k(\boldsymbol{x}^{k}_{i})}{\left \| \nabla{\cal L}_k(\boldsymbol{x}^{k}_{i}) \right \|_2},
\label{iteration}
\end{equation}
where $\nabla{\cal L}_k(\boldsymbol{x}^{k}_{i})$ refers to the gradient of ${\cal L}_k(\boldsymbol{x}^{k}_{i})$.
The step size of the Moosavi-Dezfooli et al.~\cite{DeepFool} is abnormally large due to a small gradient. To prevent this, we introduce a learning rate $\eta$ which is used together with the loss function to control the step size.
Note that the loss is large at the initial point and small near the decision boundary. 
In addition, $\epsilon$ derives the sample to cross the decision boundary as shown in the following.

If we derive the first-order Taylor series approximation of ${\cal L}_k(\boldsymbol{x}^{k}_{i+1})$ at $\boldsymbol{x}^{k}_{i}$ and substitute (\ref{iteration}) to remove $(\boldsymbol{x}^{k}_{i+1} - \boldsymbol{x}^{k}_{i})$, then we have
\begin{align}
 {\cal L}_k(\boldsymbol{x}^{k}_{i+1}) \approx &
 {\cal L}_k(\boldsymbol{x}^{k}_{i}) \left (1 - \eta \left \| \nabla{\cal L}_k(\boldsymbol{x}^{k}_{i}) \right \| _2 \right ) \nonumber \\ 
&- \eta \epsilon \left \| \nabla{\cal L}_k(\boldsymbol{x}^{k}_{i}) \right \|_2.
\label{firstorder}
\end{align}
Let us assume that we have chosen a small enough $\eta$ so that $\eta \left \| \nabla{\cal L}_k(\boldsymbol{x}^{k}_{i}) \right \|_2 < 1$. Then, if the sample approaches a decision boundary, 
${\cal L}_k(\boldsymbol{x}^{k}_{i})$ becomes small. In this case, without the last term in (\ref{firstorder}) that exists due to the introduction of $\epsilon$ in (\ref{iteration}), the loss converges to zero but does not become negative which means the sample does not cross the decision boundary. By introducing $\epsilon$ in (\ref{iteration}), the loss can become negative due to the second term in (\ref{firstorder}).

To lead the adversarial sample to a location near the decision boundary, we establish the stop conditions. The iteration stops if one of the following conditions occurs: 
\begin{eqnarray*}
(a)&& {\cal L}_k(\boldsymbol{x}^{k}_{i+1}) < 0 ~ \text{and}~ {\cal L}_k(\boldsymbol{x}^{k}_{i}) > 0 ~ ,\\
(b)&& \text{There exists any} ~\bar{k}~ \text{such that} \\
&&f_{\bar{k}} (\boldsymbol{x}^{k}_{i+1}) > \max(f_b(\boldsymbol{x}^{k}_{i+1}), f_k(\boldsymbol{x}^{k}_{i+1})),\\
(c)&& i+1 \geq I_{max},
\label{stops}
\end{eqnarray*}
where $I_{max}$ is a predefined number of maximum iterations.
Condition (a) means that the adversarial sample crosses the decision boundary. If (a) is satisfied, then the attack is successful and the resulting sample is regarded as an BSS. On the other hand, conditions (b) and (c) are about failure cases and we discard the sample if one of them is satisfied. Condition (b) means that the sample has stepped into a class that is not the target. This case occurs when there exists a non-target class between the base class and the target class. Condition (c) happens if the decision boundary is too far from the base sample.

\subsection{Knowledge distillation using BSS}
\label{sec:distill}

As mentioned in the previous section, BSSs of a teacher are beneficial for improving the generalization performance of a student classifier.
In this section, we present a method to enhance knowledge distillation by transferring information on decision boundary more precisely using BSSs.

\subsubsection{Loss function for BSS distillation}

From a training batch, our distillation scheme uses a set of base sample pairs  $\{(\boldsymbol{x}_n, c_n)~|~n=1, \cdots, N\}$ where $c_n$ denotes the class index of $\boldsymbol{x}_n$.
A set of BSSs is denoted by  $\{\boldsymbol{\mathring {x}}_n^k~|~n=1, \cdots, N; k=1, \cdots, K\}$. Let the teacher and the student classifiers be denoted by $f_t$ and $f_s$ respectively. For a sample $\boldsymbol{x_n}$, the class probability vectors are denoted  by $\boldsymbol{q}^t_n=\sigma\left(f_t(\boldsymbol{x}_n)\right)$ and $\boldsymbol{q}^s_n=\sigma\left(f_s(\boldsymbol{x}_n)\right)$, where $\sigma(\cdot)$ refers to the softmax function.
The desired class label for $\boldsymbol{x}$ is denoted by a one-hot label vector $\boldsymbol{y}^{true}$ of which the element is either one for the ground-truth class or zero for the other classes.
The proposed loss function to train the student classifier combines three losses; a classification loss $\mathcal{L}_{cls}$, the knowledge distillation loss $\mathcal{L}_{KD}$ in ~\cite{Hinton}, and an boundary supporting loss $\mathcal{L}_{BS}$:
\begin{equation}
\mathcal{L}(n) = \mathcal{L}_{cls}(n) + \alpha \mathcal{L}_{KD}(n) + \beta \sum_k^K p^k_n \mathcal{L}_{BS}(n,k).
\label{loss}
\end{equation}
If we define the entropy function as $J(\boldsymbol{a}, \boldsymbol{b}) = - \boldsymbol{a}^T \log \boldsymbol{b}$, where $\boldsymbol{a}$ and $\boldsymbol{b}$ are column vectors and $\log$ is the element-wise logarithm,
each loss is defined by
\begin{align}
\mathcal{L}_{cls}(n) &= J(\boldsymbol{y}_n^{true}, \sigma \left(f_s(\boldsymbol{x}_n)\right) ),\\
\mathcal{L}_{KD}(n) &= J(\sigma \left(\frac{f_\text{t}(\boldsymbol{{x}}_n)}{T}\right), \sigma \left(\frac{ f_s(\boldsymbol{x}_n)}{T} \right)),\\
\mathcal{L}_{BS}(n,k) &= J(\sigma \left(\frac{f_\text{t}(\boldsymbol{\mathring {x}}_n^{k})}{T}\right), \sigma \left(\frac{ f_s(\boldsymbol{\mathring {x}}_n^{k})}{T}\right)).
\end{align}

Note that $T$, the `temperature', is a design parameter to prevent the loss from becoming too large ~\cite{Hinton}.
$p^k_n$ in the third term of (\ref{loss}) is the probability of class $k$ being selected as the target class, which is introduced to sample target classes stochastically during training.
The definition of $p^k_n$ can be found in (\ref{target_sample}).
The linearly decaying weights are used for $\alpha$ and $\beta$, following the common practice in existing knowledge distillation~\cite{FITNET,Attention} techniques.
Note that the $\mathcal{L}_{cls}$ transfers direct answers (one-hot labels) for the training samples, whereas $\mathcal{L}_{KD}$ transfers probabilistic labels~\cite{Hinton}.
In contrast, the boundary supporting loss $\mathcal{L}_{BS}$ is introduced to transfer the information about the decision boundary directly.

\subsection{Miscellaneous issues on using BSSs}
\label{issues_BSS}

\subsubsection{Base sample selection for boundary supporting loss.}
To reduce the computation, we select $N$ base samples out of $N_{batch}$ training samples according to a specific rule explained below and apply the boundary supporting loss only to the selected samples.

The base samples for generating the adversarial samples are selected from the training batch
$\mathcal{B} = \{(\boldsymbol{x}_n, c_n) ~|~ n= 1,2, \cdots , N_{batch}\}$. 
A training sample pair ($\boldsymbol{x}_n, c_n$) is selected as the base sample for an adversarial attack if the class $c_n$ has the highest probability for both the teacher and the student classifiers. 
That is, considering the probability vectors $\boldsymbol{q}^t_n$ and $\boldsymbol{q}^s_n$, the base sample set is determined by
\begin{align}
\mathcal{C} = \{(\boldsymbol{x}_n,& c_n)  \,|\, \underset{c}
{\mathrm{arg max}}(\boldsymbol{q}^t_{n,c}) \equiv c_n, \nonumber \\
&\underset{c}
{\mathrm{arg max}}(\boldsymbol{q}^s_{n,c}) \equiv c_n, (\boldsymbol{x}_n, c_n)\in\mathcal{B} \}
\end{align}
where $\boldsymbol{q}_{n,c}^o$ ($o=t, s$) is the $c$th element of $\boldsymbol{q}_n^o$.
If the size of $\mathcal{C}$ is smaller than a predefined $N$, all the samples in $\mathcal{C}$ is used for the boundary supporting loss.
If the size of $\mathcal{C}$ is larger than $N$, we select $N$ samples that have the highest distance between $\boldsymbol{q}^t_n$ and $\boldsymbol{q}^s_n$, i.e.,
\begin{equation}
d_n = \left \| \boldsymbol{q}^t_n - \boldsymbol{q}^s_n \right \|_2^2.
\label{selection_measure}
\end{equation}
Large $d_n$ means that the probability vector $\boldsymbol{q}^t_n$ of the teacher and the probability vector $\boldsymbol{q}^s_n$ of the student are largely different from each other at the base sample position $\boldsymbol{x}_n$.
Since the reduction of $d_n$  matches the goal of knowledge distillation, it is reasonable to choose a base sample with large $d_n$.

\subsubsection{Target class sampling.} 
A BSS can target all classes except the base class.
In the learning process, one of the classes is selected as the target class according to the following criteria and an BSS is generated by adversarial attacking to the selected target class.
For the base sample $\boldsymbol{x}_n$, the probability $p^k_n$ to sample the class $k$ is defined based on the class probability $\boldsymbol{q}^t_n$ of the teacher as follows:
\begin{equation}
p^k_n = 
\begin{cases}
0 ,& \text{if } k = c_n\\
\boldsymbol{q}^t_{n,k}/(1 - \boldsymbol{q}^t_{n, c_n}),              & \text{otherwise}.
\end{cases}
\label{target_sample}
\end{equation}

This is motivated from that it is important to precisely transfer the knowledge on the decision boundary between two classes that are hard to discriminate from each other. $\boldsymbol{q}^t_{n,k\neq c_n}$ 
having high value means that the class $k$ is hard to discriminate from the base class $c_n$ for the base sample $\boldsymbol{x}_n$. Therefore, the target class is sampled with priority given to the class with a high $\boldsymbol{q}^t_{n,(\cdot)}$ for  $\boldsymbol{x}_n$.

\begin{table*}[t]
\centering
\caption{Comparison on CIFAR-10 dataset}
\label{CIFAR10}
\begin{tabular}{@{}cccccccc@{}}
\toprule
Student & Original & \begin{tabular}[c]{@{}c@{}}Hinton\\ \shortcite{Hinton}\end{tabular}  & \begin{tabular}[c]{@{}c@{}}FITNET~\shortcite{FITNET}\\ +Hinton\end{tabular} & \begin{tabular}[c]{@{}c@{}}AT~\shortcite{Attention}\\ +Hinton\end{tabular} & \begin{tabular}[c]{@{}c@{}}FSP~\shortcite{Gift_distill}\\ +Hinton\end{tabular} & Proposed & \begin{tabular}[c]{@{}c@{}}FSP~\shortcite{Gift_distill}\\ +Proposed\end{tabular} \\ \midrule
ResNet 8         & 86.02\%  & 86.66\% & 86.73\%                                                  & 86.86\%                                              & 87.07\%                                               & \underline{87.32\%}  & \textbf{87.52\%}                                                 \\
ResNet 14        & 89.11\%  & 89.75\% & 89.72\%                                                  & 89.84\%                                              & 89.92\%                                                & \textbf{90.34\%}  & \underline{90.13\%}                                          \\
ResNet 20        & 90.16\%  & 90.77\% & 90.46\%                                                 & \underline{90.81\%}                      &  90.27\%                                              & \textbf{91.23\%}  & 90.19\%                                                 \\ \bottomrule
\end{tabular}
% \vskip -0.1in
\end{table*}

\subsection{Metrics for similarity of decision boundaries}\label{sec:metric}

To verify whether the proposed method actually transfers decision boundaries in knowledge distillation, we need some metrics. Here, we propose two metrics based on BSSs to measure the similarity between the decision boundaries of two classifiers (i.e., teacher and student classifiers in knowledge distillation). These metrics are used to evaluate the performance of knowledge distillation or analyze the benefits of BSSs in knowledge distillation.

Given the $n$th base sample $\boldsymbol{x}_n$,
the perturbation vector to attack the target class $k$ for the teacher classifier is obtained by 
\begin{equation}
\boldsymbol{\bar{x}}^{k,t}_n = \boldsymbol{\mathring{x}}^{k,t}_n - \boldsymbol{x}_n.
\end{equation}
Likewise, $\boldsymbol{\bar{x}}^{k,s}_n$ denotes the perturbation vector for the student classifier. 
Using a set of perturbation vector pairs $\{ (\boldsymbol{\bar{x}}^{k,t}_n, \boldsymbol{\bar{x}}^{k,s}_n)~ |~ n=1,2, \cdots, N; k=1, \cdots, K\}$,
the similarity between the two decision boundaries is defined by two metrics: The {\it Magnitude Similarity (MagSim)} in (\ref{metric1}) and the {\it Angle similarity (AngSim)} in (\ref{metric2}):
\begin{align}
MagSim&=\frac{1}{N {K}} \sum_{n=1}^N \sum_{k=1}^{K} \frac{\min(\norm{\boldsymbol{\bar{x}}_n^{k,t}}_2, \norm{\boldsymbol{\bar{x}}_n^{k,s}}_2)}{\max(\norm{\boldsymbol{\bar{x}}_n^{k,t}}_2, \norm{\boldsymbol{\bar{x}}_n^{k,s}}_2)} \label{metric1} \\
AngSim&=\frac{1}{N {K}} \sum_{n=1}^N 
\sum_{k=1}^{K}
\frac{<\boldsymbol{\bar{x}}_n^{k,t},\boldsymbol{\bar{x}}_n^{k,s}>}
{\norm{ \boldsymbol{\bar{x}}_n^{k,t} }_2 \times \norm{ \boldsymbol{\bar{x}}_n^{k,s} }_2}.
\label{metric2}
\end{align}
These two metrics have values in the range of [0,1] and higher values represent more similar decision boundaries.

Note that {\it MagSim} represents the similarity with respect to the distance from the base sample to the decision boundary.
and {\it AngSim} depicts that with respect to the path direction from the base sample to the boundary.
Since the path is obtained by the gradient of the classification score function, {\it AngSim} reflects the similarity with respect to the surface shape of the class score function which affects the shape of decision boundary. Hence, we can say that the decision boundaries of two classifiers have become more similar if either of the metrics increases.

\section{Experiments}
\label{sec:exp}

Through experiments, we show that the proposed method is a way to enhance the performance of knowledge distillation.
In order to increase the reliability of the experiment, we performed the training 10 times for the same condition, and displayed the average of the results.
Experiments were performed on the CIFAR-10~\cite{CIFAR}, ImageNet 32$\times$32~\cite{ImageNet32} and TinyImageNet datasets using a variety of residual networks~\cite{resnet}.

\subsection{Performance on image classification}
The performance of the proposed method is verified by image classification on the CIFAR-10, ImageNet 32$\times$32, and TinyImageNet datasets.
We trained the student classifiers in seven different ways.
The first method is denoted as `original', which uses only the classification loss for training.
The second method is denoted as `Hinton'.
Using the classification loss and the KD loss, 
`Hinton' was implemented in the same way as in Hinton et al.~\cite{Hinton}.
The next three methods are the latest knowledge distillation methods implemented with the KD loss.
The `FITNET'~\cite{FITNET} transfers the response of the intermediate layer.
The method denoted as `AT' is transferring the spatial attention of the teacher classifier to the student classifier~\cite{Attention}.
The `FSP' simplifies the layer response of the teacher into a channel-wise correlation matrix, which is used as the medium of knowledge transfer~\cite{Gift_distill}.
Since the three methods use the KD loss, they are labeled together with `+ Hinton'.
The last two methods are; the proposed method which is denoted as `proposed' and the proposed method implemented together with the `FSP' method which is denoted as `FSP+proposed'.
The performance of all classifiers was measured in terms of accuracy.

% Please add the following required packages to your document preamble:
% \usepackage{booktabs}
\begin{table*}[t]
\centering
\caption{Comparison on ImageNet 32$\times$32}
\label{ImageNet32}
\begin{tabular}{@{}cccccccc@{}}
\toprule
     & Original & \begin{tabular}[c]{@{}c@{}}Hinton\\ \shortcite{Hinton}\end{tabular}  & \begin{tabular}[c]{@{}c@{}}FITNET~\shortcite{FITNET}\\ +Hinton\end{tabular} & \begin{tabular}[c]{@{}c@{}}AT~\shortcite{Attention}\\ +Hinton\end{tabular} & \begin{tabular}[c]{@{}c@{}}FSP~\shortcite{Gift_distill}\\ +Hinton\end{tabular} & Proposed & \begin{tabular}[c]{@{}c@{}}FSP~\shortcite{Gift_distill}\\ +Proposed\end{tabular} \\ \midrule
% & Baseline & Hinton  & FITNET  & AT      & FSP     & Proposed & FSP+Proposed \\ \midrule
Top1 acc & 31.94\%  & 32.43\% & 32.60\% & 32.61\% & 32.66\% & \underline{32.69\%}  & \bf{32.72\%}      \\
Top5 acc & 56.21\%  & 56.99\% & 57.02\% & 57.14\% & 57.14\% & \underline{57.17\%}  & \bf{57.27\%}      \\ \bottomrule
\end{tabular}
% \vskip -0.1in
\end{table*}

\begin{table*}[t]
\centering
\caption{Comparison on TinyImageNet}
\label{TinyImageNet}
\begin{tabular}{@{}cccccccc@{}}
\toprule
     & Original & \begin{tabular}[c]{@{}c@{}}Hinton\\ \shortcite{Hinton}\end{tabular}  & \begin{tabular}[c]{@{}c@{}}FITNET~\shortcite{FITNET}\\ +Hinton\end{tabular} & \begin{tabular}[c]{@{}c@{}}AT~\shortcite{Attention}\\ +Hinton\end{tabular} & \begin{tabular}[c]{@{}c@{}}FSP~\shortcite{Gift_distill}\\ +Hinton\end{tabular} & Proposed & \begin{tabular}[c]{@{}c@{}}FSP~\shortcite{Gift_distill}\\ +Proposed\end{tabular} \\ \midrule
% & Baseline & Hinton  & FITNET  & AT      & FSP     & Proposed & FSP+Proposed \\ \midrule
Top1 acc & 50.68\%  & 52.35\% & \underline{53.52\%} & 52.74\% & 53.43\% & 52.99\%  & \bf{53.86\%}      \\
Top5 acc & 76.14\%  & 77.67\% & 78.10\% & 77.82\% & 78.15\% & \underline{78.38\%}  & \bf{78.49\%}      \\ \bottomrule
\end{tabular}
% \vskip -0.1in
\end{table*}

\subsubsection{CIFAR-10.} 
The CIFAR-10 is a classification dataset with 10 classes and 32x32 resolution, consisting of 50k training images and 10k test images.
ResNet26 with 92.55\% accuracy is used as a teacher classifier.
Meanwhile, ResNet8, ResNet14 and ResNet20 are used as student classifiers.
All classifiers were trained over 80 epochs. `FITNET' and `FSP', which require two-stage learning scheme, were learned over 80 epochs after using 40 epochs for initialization.
Table~\ref{CIFAR10} shows the result. 
The proposed method shows improved performance compared 
to Hinton.
Also, the performance improvement of the proposed method is better than the existing state-of-the-arts.
This confirms that the additional samples of the proposed method is useful for the knowledge distillation.

\subsubsection{ImageNet 32$\times$32.}
ImageNet 32$\times$32 is a 32$\times$32-downsampled version of the ImageNet dataset~\cite{ImageNet}
This dataset is a classification dataset with 1000 classes, consisting of 1,281k training images and 50k validation images.
ResNet32 with 48.04\% top-1 accuracy and 73.22\% top-5 accuracy is used as a teacher classifier and ResNet8 is used as a student classifier.
All classifiers were trained over 40 epochs. `FITNET' and `FSP' spent 4 epochs for initialization.
Table~\ref{ImageNet32} shows the result.
The proposed method shows better performance than  Hinton method and shows comparable performance with other state-of-the-arts.
When combined with `FSP', the proposed method shows better performance for top-5 accuracy.

\subsubsection{TinyImageNet.} TinyImageNet is a subset of the ImageNet dataset with 64$\times$64 resolution.
It contains 100k training images and 10k test images in 200 classes.
ResNet42 is used for the teacher classifier, which has 56.10\% top-1 accuracy and 78.71\% top-5 accuracy.
ResNet10 is selected for the student classifier.
Classifiers were trained for 80 epochs. 
`FITNET' and `FSP' spent 10 epochs for initialization.
The result is described in Table~\ref{TinyImageNet}.
The proposed method shows better results than Hinton and `AT'.
Although the proposed method shows lower top-1 accuracy than `FITNET' and `FSP' which require additional learning steps, it has higher top-5 accuracy than other state-of-the-arts.
Also, the proposed method shows performance superior to those of other algorithms when combined with `FSP'.

% Please add the following required packages to your document preamble:
% \usepackage{booktabs}
\begin{table*}[t]
\centering
\caption{Comparison of knowledge distillation using various types of adversarial samples.}
\label{attack_compare}
\begin{tabular}{@{}ccccccc@{}}
\toprule
  & Baseline & Random noise & L2 minimize & FGSM~\shortcite{GoodFellow} & DeepFool~\shortcite{DeepFool} & Proposed \\ \midrule
Accuracy & 86.66\%  & 86.73\%      & 86.94\%     & \underline{87.06\%} & 86.95\%  & \textbf{87.32\%}  \\ \bottomrule
\end{tabular}
% \vskip -0.2in
\end{table*}

\begin{figure*}[t]
\begin{center}
\centerline{\includegraphics[width=0.9\linewidth]{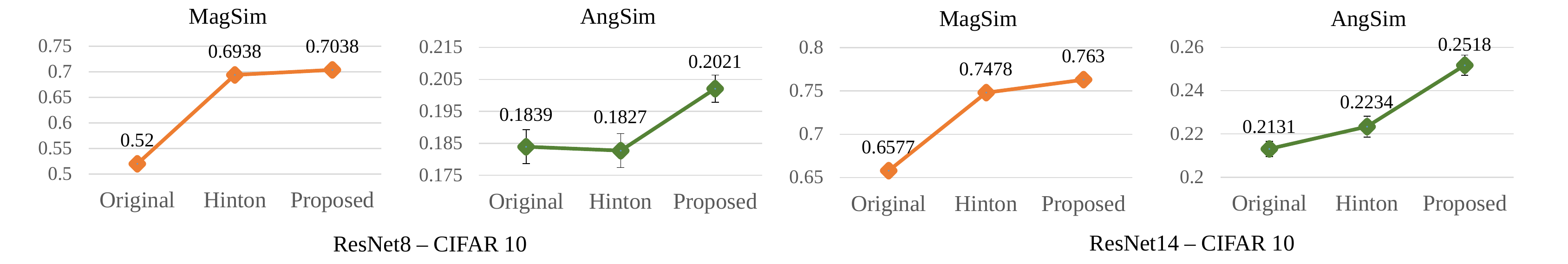}}
\caption{Evaluation of proposed method for decision boundary similarities (\textit{MagSim}, \textit{AngSim}).}
\label{figure_exp2}
\end{center}
\end{figure*}

\subsection{Generalization of the classifier}

\begin{figure}[t]
\begin{center}
\centerline{\includegraphics[width=1.0\linewidth]{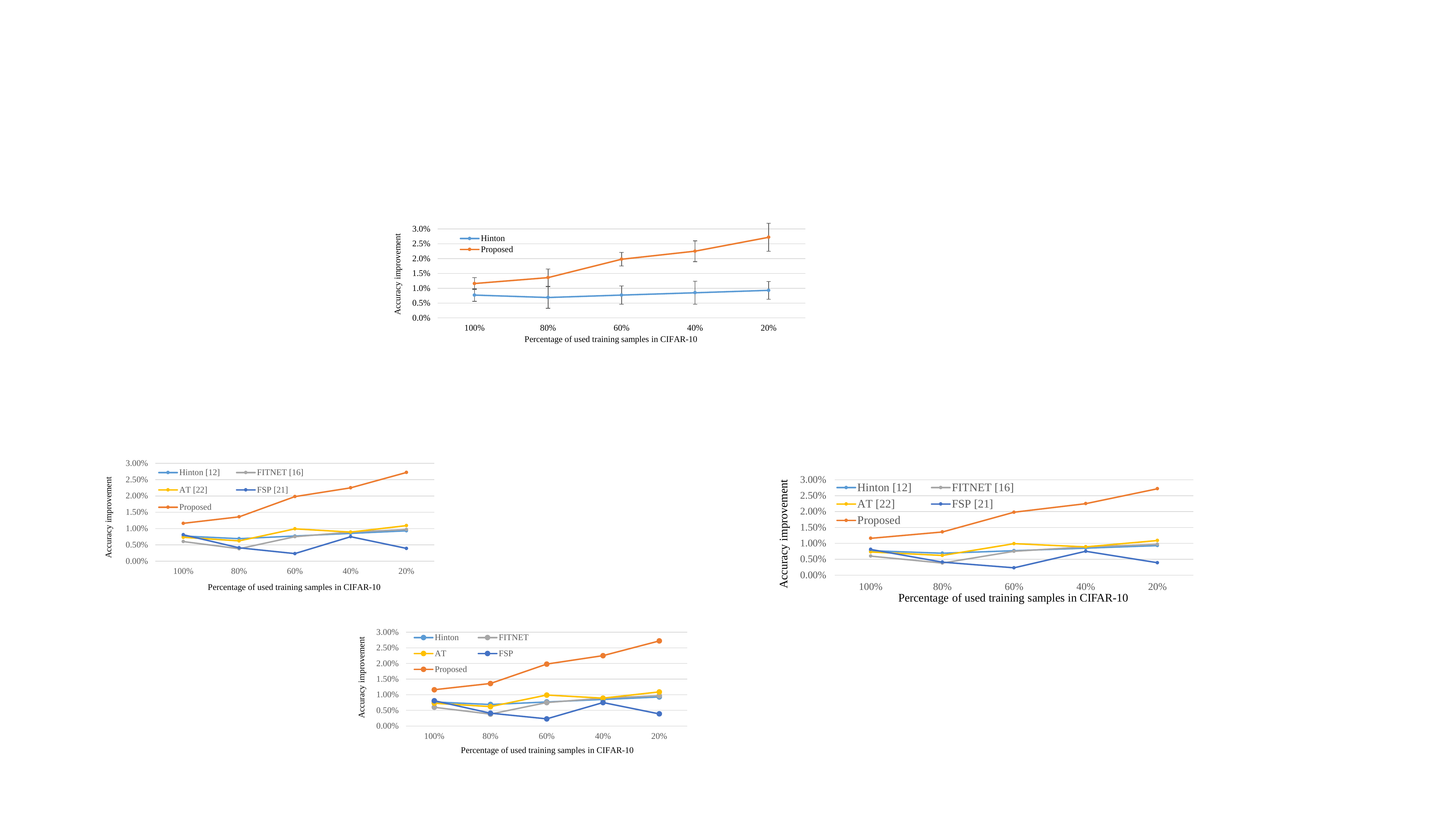}}
\caption{Generalization of the classifier. The smaller the number of training samples are, the larger improvement the proposed method shows.}
\label{figure_exp3}
\end{center}
\end{figure} 

The proposed method improves the generalization performance of a student classifier using the samples supporting the decision boundary obtained from a teacher classifier.
Through an experiment, we verified that the generalization performance of the student classifier actually increases by the proposed method.
In order to measure the generalization performance, the experiment was repeated while reducing the number of training samples from 100\% to 20\%. The CIFAR-10 dataset was used in this experiment, and
ResNet26 trained on the whole dataset was used as the teacher classifier while ResNet14 was used as the student classifier.
All methods were trained on the same training data for fairness.

Figure~\ref{figure_exp3} shows the performance improvement from the original method for the size of the dataset.
Here, we can see that the performance improvement of the other methods does not change much regardless of the size of data. 
On the other hand, the proposed method shows bigger performance improvement for less training data.
In a situation where it is difficult to achieve generalization due to insufficient data, the proposed method shows a large performance improvement, which means that the proposed method improves the generalization of the student classifier.

\subsection{Analysis with similarity measure}

We conducted an experiment to analyze the effect of the proposed method on the decision boundary of the network.
The experiment is to measure the similarity metrics (\textit{MagSim}, \textit{AngSim}) of trained student and teacher.
Two similarity metrics reflect the similarity of decision boundaries.
Thus, high similarity metrics mean that decision boundary is transferred by knowledge distillation.
Experiment was performed in a CIFAR-10 test set.
ResNet 8 and Resnet 14 were used.
`Original', `Hinton', and the proposed method were tested.
The experimental results are shown in Figure~\ref{figure_exp2}.
Compared with `original', the Hinton method mainly increases \textit{MagSim} and \textit{AngSim} changes are small.
In other words, the Hinton method transfers only the distance to the decision boundary and does not consider the direction.
On the other hand, the proposed method increases both \textit{MagSim} and \textit{AngSim} when compared to 'original'.
Therefore, the proposed method transfers both distance and direction of decision boundary.
The experiment show that the proposed method transfers the decision boundary more accurately and explains the reason for the high performance in the previous experiments.

\subsection{Self-comparison}
We conducted self-comparisons to analyze the effects of a boundary supporting sample and miscellaneous issues.
The experiments were performed on the CIFAR-10 dataset.
ResNet26 was used as the teacher classifier, and ResNet8 was the student classifier.
A boundary supporting sample (BSS) is an adversarial sample that especially designed to reflect the information about a decision boundary.
Therefore, a BSS is more suitable for knowledge distillation than other types of adversarial samples.
To verify this, we tested different kinds of adversarial attacks for knowledge distillation.
Experiments were performed on five kinds of adversarial samples.
The results are described in Table~\ref{attack_compare}.
The `Baseline' shows the performance of knowledge distillation without the proposed boundary supporting loss.
The `Random noise' uses a randomly generated noise instead of a gradient-based adversarial sample.
The method denoted as `L2 minimize' presents the performance of the proposed method with adversarial samples calculated based on the L2 minimization of (\ref{score_diff}).
The `FGSM'~\cite{GoodFellow} and `DeepFool'~\cite{DeepFool} use other well-known attack methods for the proposed method.
The `Proposed' is the proposed distillation method with BSS.
The result shows that all adversarial samples including random noise are beneficial to knowledge distillation.
Random noise shows the smallest performance improvement.
The gradient-based methods except the proposed method show similar performance improvement.
On the other hand, the proposed method using BSS shows the greatest improvement, showing that a BSS is more suitable for knowledge distillation.

Experiment on miscellaneous issues is presented in Table~\ref{ablation}.
'Proposed' shows the performance of the proposed method using base sample selection and target class sampling.
'All selection' shows performance when all samples are used as base samples, and 'Random selection' shows performance when base samples are randomly selected without the proposed scheme.
Two results show that the proposed base sample selection not only reduces computation but also contributes to performance.
The performance when the target class is selected randomly without the proposed sampling is shown in 'Random target class'.
The result implies that the proposed sampling according to the class probability is reasonable and effective method.

\subsection{Implementation details}
All the experiments were performed using residual networks~\cite{resnet}.
The channel sizes of ResNet were set to 16, 32, and 64 for CIFAR-10, and 32, 64, and 128 for ImageNet 32$\times$32.
For TinyImageNet, ResNet with four block was used with channel size of 16, 32, 64, and 128.
We used random crop and random horizontal flip for data augmentation and normalized an input image based on the mean and the variance of the dataset.
The temperatures of the KD loss and the adversarial loss were fixed to 3 in all experiments.
The parameter $\alpha$ in (\ref{loss}) was initialized to 4 and linearly decreased to 1 at the end of training.
The $\beta$ in (\ref{loss}) was set to 2 initially and linearly decreased to 0 at the 75\% of the whole training procedure, based on our empirical observations: When $\beta$ was not zero at the final training stage, the performance was degraded.
The learning process was performed with 256 batch size, with a learning rate which started at 0.1 and decreased to 0.01 at half of the maximum epoch and to 0.001 in 3/4 of the maximum epoch.
The momentum used in the study was 0.9 and the weight decay was 0.0001.
$\eta=0.3$ was used for the adversarial attack in the proposed method and the maximum number of iteration was set to 10 for knowledge distillation.
For the boundary supporting loss, $N_{adv}=64$ was selected among 256 batch samples.

\begin{table}[t]
\centering
\caption{Self-comparison on base sample selection and target class selection.}
\label{ablation}
\begin{tabular}{@{}ccccc@{}}
\toprule
Proposed & \begin{tabular}[c]{@{}c@{}}All sample\\ selection\end{tabular} & \begin{tabular}[c]{@{}c@{}}Random \\ selection\end{tabular} & \begin{tabular}[c]{@{}c@{}}Random\\ target class\end{tabular} & Original \\ \midrule
87.32\%  & 87.18\%                                                        & 87.10\%                                                           & 87.08\%                                                       & 86.02\%  \\ \bottomrule
\end{tabular}
\end{table}

\section{Conclusion}
\label{con}

In this paper, we have investigated informative samples for efficient knowledge transfer.
The adversarial attack method was modified to find a boundary supporting sample (BSS) supporting a decision boundary.
Based on the BSS, we proposed a knowledge distillation method to transfer more accurate information about the decision boundary.
Experiments have shown that the proposed method improves the performance of knowledge distillation.
Also, it was shown that the proposed method has stronger generalization performance and so it is more effective in situations with fewer training samples.
Designing a knowledge distillation method in terms of sample manipulation is a new direction that has not been attempted in the past studies.
It is also a new approach to utilize an adversarial attack to find and transfer the information about the decision boundary.
Therefore, this work can be useful for future research on knowledge distillation and on the application of an adversarial attack.

\section{Acknowledgement}

This work was supported by  
Next-Generation ICD Program through NRF funded by Ministry of S\&ICT [2017M3C4A7077582] and
ICT R\&D program of MSIP/IITP [No.B0101-15-0552, Predictive Visual Intelligence Technology].

\bibliographystyle{aaai}
\bibliography{refs}

\end{document}